\documentclass[a4paper]{article}
\usepackage{pdfsync}
\usepackage[utf8]{inputenc}
\usepackage{graphicx}
\usepackage{amssymb,amsfonts,amsmath,amsthm}
\usepackage{hyperref}
\usepackage{multirow}
\usepackage{verbatim}
\usepackage{xcolor}
\usepackage{booktabs}
\usepackage[sectionbib,round]{natbib}
\usepackage{textcomp}
\usepackage{upquote}
\usepackage{bm}
\usepackage{dirtree}
\usepackage{algorithm}%
\usepackage{algorithmicx}%
\usepackage{algpseudocode}%

\theoremstyle{plain}

\theoremstyle{definition}

\begin{document}

\title{\textbf{Applying Data Driven Decision Making to rank Vocational and Educational Training Programs with TOPSIS}}

%\author{Author
%\\{\small Affiliation}}

\author{J.M. Conejero$^1$, J.C. Preciado$^1$, A.E. Prieto$^1$, M.C. Bas$^2$, V.J. Bol\'os$^2$ \\ \\
{\small $^1$ Dpto. Ingenier\'{\i}a Sistemas Inform\'aticos y Telem\'aticos.} \\
{\small Universidad de Extremadura. Avda. de la Universidad, 10071 C\'aceres, Spain.} \\
{\small $^2$ Dpto. Matem\'aticas para la Econom\'{\i}a y la Empresa, Facultad de Econom\'{\i}a.} \\
{\small Universitat de Val\`encia. Avda. Tarongers s/n, 46022 Valencia, Spain.} \\
{\small e-mail\textup{: \texttt{chemacm@unex.es}, \texttt{jcpreciado@unex.es}, \texttt{aeprieto@unex.es} }} \\
{\small \texttt{maria.c.bas@uv.es}, \texttt{vicente.bolos@uv.es}}
}

\date{March 2021}

\maketitle

\begin{abstract}
In this paper we present a multi-criteria classification of Vocational and Educational Programs in Extremadura (Spain) during the period 2009-2016. This ranking has been carried out through the integration into a complete database of the detailed information of individuals finishing such studies together with their labor data. The multicriteria method used is TOPSIS together with a new decision support method for assessing the influence of each criterion and its dependence on the weights assigned to them. This new method is based on a worst-best case scenario analysis and it is compared to a well known global sensitivity analysis technique based on the Pearson's correlation ratio.
\end{abstract}

\section{Introduction}
The 2008 financial crisis that hit the world\textsc{\char13}s economies has had a particularly acute impact in Spain \citep{guardiola2015income}. It is only since 2014 that Spain seemed to begin its recovery \citep{marti2015}. However, this recuperation is still far to be acceptable with regard to the labor landscape \citep{casares2018}.

One of the main Spanish weaknesses that the crisis exposed was the so-called duality of the labor market. Thus, Spain is characterized by the existence of two very different types of workers. On one hand, long term workers on indefinite contracts, having both a very high job security and a very high cost for companies (especially in terms of dismissals) and usually with university studies even for jobs that do not require them. On the other hand, short term workers on temporary contracts or seasonal contracts with low wages and, in most cases, with very little training. 

Another structural weakness of the Spanish economy unveiled during the years of the crisis was the fact that it had been relied heavily on two pillars: construction and tourism (and their associated services). This productive model had its main Achilles heel in the low level studies required in many of the jobs created in both sectors. Moreover, the relatively high wages that a worker could earn before the crisis, mainly in construction, led many young people to abandon their studies to work in these industries, without prior quality training. When the crisis arose and the destruction of employment reached unprecedented levels, Spain discovered that had to deal with a mass of unemployed, mostly young, people who, having no adequate training, had very difficult or even impossible reinstatement into the labor market.

This problem has been most pronounced in some regions of Spain as Extremadura. Extremadura is a European Union Objective 1 region located in western Spain that according to the Eurostat Regional Yearbook 2018\footnote{https://cutt.ly/as19ww2}, its GDP per inhabitant in relation to the EU-28 average is 61.47\%, it has 23.7\% of unemployment rate and, even worse, its early leavers from education and training of young people rate is 20.9\% and its young people neither in employment nor in education or training rate is 20\%. 

In other countries where the economic model was more diversified, with large sectors of skilled employment and better trained workers, the crisis was less intense, the employment destruction less acute, and the recovery was faster.
One of the differences between Spain and, particularly, Extremadura with respect to those countries is the importance they give to Vocational Eucation and Training (hereinafter VET). For the European Union, VET should ``prepare young people for entering and successfully and sustainably participating in the labor markets as well as to enable high potentials (e.g. migrants, refugees, low-skilled and unemployed, inactive groups, including women) to stay and/or (re-)enter the labor market'' \citep{AdvisoryCommitteeonVocationalTraining2018}. In Germany, for example, VET studies are closely linked to the labor market, so that the majority of VET students are also trained in companies where, in many cases, they end up working. This means that there is a quarry of workers with a specific qualification for the needs of the labor market.

However, in the collective imagination of Spanish families, VET has been considered for decades  a second-rate training and has not been much appreciated. This vision, together with the high drop-out rates, has caused a very marked duality in training: on the one hand people who either did not finish more than compulsory education or, at best, have VET studies (which in this last case are seen wrongly as a low level education), or people with university education who, due to the high unemployment rates suffered in Spain during the crisis (still persisting), are hired in positions for which such studies are not really required, causing another of the many big problems of the Spanish labor market, which is the overqualification of workers \citep{flisi2017measuring}.

In order to try to alleviate the mentioned problems, the Government of Extremadura, providing its historical data, asked for a scientific analysis about the real impact of VET studies on accessing to the labor market with a two-fold goal: increase the resources of those VET studies with higher employment rate and promote such studies among their unemployed citizens enhancing the image of the VET studies that really help to get a job.

To this end, the aim of this work is to determine the efficiency of VET studies and the evaluation of the performance of VET graduates from Extremadura in the different degrees of VET programs in the labor market. Thus, we illustrate a data-driven multi-criteria decision-making methodology with the aim of classifying the different degrees of VET programs according to some criteria related to labor insertion. Concretely, we have applied TOPSIS (Technique for Order Performance by Similarity to Ideal Solution) \citep{Hwang1981} to this problem. TOPSIS is a well-known classical MCDM method, widely used by researchers and practitioners, that supports decision makers in performing analysis, comparisons and rankings to select the best alternative using a finite number of criteria. Moreover, since all criteria are weighted, their importance may be modified providing, thus, the flexibility for creating different rankings prioritizing different aspects. In our case, we have considered 8 different criteria whose values have been computed from a real dataset with more than 28000 VET student records containing both educational and labor information. This is a key contribution for this work since the information obtained is enriched by real data instead of being based on questionnaires as similar TOPSIS approaches \citep{Rad2011} or the REFLEX project in the field of higher education \citep{Allen2006}. Furthermore, to the best of our knowledge, the relationship between VET programs and labor market has not been  explored by previous works with such level of detail.

The rest of the article is organized as follows. Section 2 reviews some works related with predicting different outcomes using academic data, and some other works using TOPSIS as Multi-Criteria Decision Analysis method together with a weight assignment analysis, on different scientific domains. Section 3 describes the datasets used and the process applied to them for computing the different data used in our study. Section 4 describes the methodology followed to apply TOPSIS. Section 5 explains the influence of the criteria applied during the process. The results obtained and further considerations are detailed in Section 6. Finally, Section 7 concludes the paper.

\section{Literature review}

The analysis of educational data is considered today as one of the foundations to implement new educational policies and may be even more so in the future \citep{Williamson2016}.

In this sense, there are several works focused on analysing success or failure in the pre-university stages. Thus, one example of this topic is the work of\cite{Sen2012} that tries to predict the outcome of Turkish high school students in the examinations of national selection that these students must perform. 
Another example is the work of \cite{Sara2015} where they analyze the data of more than 36.000 Danish students who, at least, have completed the first six months in secondary education institutes, to predict if they were going to abandon their studies in the next 3 months.
\cite{Aguiar2015} analyzed 11.000 American students from different high school courses to try to detect who were at risk of abandonment and try to take appropriate measures so that it does not occur. 
%There are more works as well that, like the previous one, try to predict dropout during high school studies \cite{Marquez-Vera2016} \cite{Sansone2019}.

There are also a number of works within this field focused on university students. 
One example is the work of \cite{Campagni2015}, in which the authors analyze the academic data to define an ideal trajectory for university students and compare real students based on that ideal trajectory. 
\cite{DiPietroEducation} analyze the academic performance in terms of student performance, teaching activities and student satisfaction among others.
Continuing toward the goal of predicting the student performance in the university, \cite{AsifEducation} make use of data mining and clustering methods to predict the graduation performance of a student given only his pre-university marks.
%Ahadi et al. \cite{Ahadi2015} analyze different data of university students to identify those who may need personalized attention.
\cite{Fernandez-Garcia2018} perform a comparison of algorithms that can predict which first-year university students are more likely to finish their studies and apply their conclusions in the students of the University of Almer\'\i{}a (Spain).
%There are another examples of more works focused on predicting university student performance \cite{Vihavainen2013} \cite{Porter2014}  \cite{Jia2015}.

More related with our work, that is, analyzing the possible relationship between the education of the students and their employability, we could start by mentioning the one of \cite{Jackson2014a} where she analyzes data of 56255 Australian Bachelor degree graduates in 2011 and 2012 to identify which factors influence graduate employment. 
Also, in Australia, \cite{Mewburn2018} perform an analysis of the extent of demand for Ph.D. student skills and capabilities in the Australian employment market. 
\cite{Bharambe2017} analyse 5 skills of 91177 Indian students to predict the likelihood of each one of getting a job.
\cite{Thakar2017a} use 151 attributes from 7143 Indian university students to conclude that only 8 attributes play a significant role in predicting students’ employability in the  first year of their enrollment. 
%Piad \cite{Piad2018} uses the data of 515 Philippines IT graduates to determine whether the IT graduates will get an IT related profession or not related and conclude that the higher the scores in IT subjects the higher the likelihood of getting a job in their field. 
\cite{Garcia-Penalvo2018} use information from 3000 Spanish students to build predictive models that define how these students get a job after finalizing their university degrees.

From the analysis of the state of the art in the domain of this project, we can observe that, although there are numerous works that use data from the educational field to try to predict possible dropouts, performance and even employability, we do not have evidence of anyone trying to analyze the possible relationship between VET programs and employment. 

Concerning the methodological aspects of the present paper, several studies in different domains have used multi-criteria decision-making methods and, in particular TOPSIS methodology, to select the best alternative using a finite number of criteria. Indeed, in many studies new extensions or improvements for the TOPSIS methodology have been proposed. For instance, \cite{yue2011,yue2012} proposes an extension of TOPSIS to determine the weights of the experts, considering a level of uncertainty assigned to the criteria. \cite{olson2004} uses different distance metrics and shows how the results of the TOPSIS methodology depend on both the weighting scheme and the distance metric used. Other study \citep{dincer2016} proposes an extension of the TOPSIS methodology (Fuzzy TOPSIS) combined with a weighting criteria methodology (Fuzzy AHP). In the last step of this method the authors apply sensitivity analysis based on the weights assigned by the experts to observe how the preferences of the decision makers would affect the final classification. To the best of our knowledge, the works that use sensitivity analysis as a final step of the TOPSIS or extended TOPSIS methodology are based on the weighting scheme assigned by the experts, regardless the weighting methodology used. The proposed decision support method (scenario comparison methodology) provides a useful tool for the experts that shows the sensitivity of each criterion to possible changes in the weighting scheme. It has the advantage of not needing a previous definition of the criteria weights by the experts, and therefore the information delivered by the method is readily available to the expert panel before they make the weight assignment.

\section{Data management}
This section presents the datasets used to establish the different rankings presented in this paper together with the whole process followed to obtain them. To this purpose, a collaboration agreement was established with the data owners, the Education and Employment Board of the Government of Extremadura, to acquire the data. Based on this agreement, we established several meetings with them in order to identify the information available and needed to perform the analysis. Then, the process for obtaining the data was divided into three different steps: i) data sources identification; ii) data gathering and iii) data warehouse design. These steps will be explained in the next subsections. 

\subsection{Data sources identification}

Since the main purpose of this study was to analyze the utility of the different VET programs in the region in terms of employability, the meetings with the Educational and Employment Board were mainly focused on identifying the data sources that provide information regarding those two areas. Fortunately, this information was acquired in the region by means of two different sources that directly depend on the aforementioned board: 
\begin{itemize}
\item Rayuela\footnote{https://cutt.ly/Ws0d1Mq} is the digital platform used by all the education centres to support the teaching staff in all the students management tasks such as exam realization, evaluation and so on. The platform includes data belonging to different education centres, concretely: Middle Schools, High Schools, VET Schools and Official Languages Schools. In this study we were mainly interested in the data from VET Schools since we wanted to analyze this kind of studies. 
\item Public Regional Employment Service (SEXPE)\footnote{https://cutt.ly/Gs0fobj} is the organization provided by the public administration to manage all the tasks related to labor in the region. This service manages any job contract signed in the region as well as the employment demands and subsidies for unemployed people in the region. Thus, all the information regarding the citizens' working life is provided by this public service.
\end{itemize}
Two additional sources of information were also used in this study:
\begin{itemize}
\item National Social Security Agency\footnote{https://cutt.ly/Xs0fELz} is a national organization that provides the public insurance to all the workers in the country. To this purpose, the Agency also recovers the labor information for the citizens independently of the region where they are working. The information provided was useful for complementing the data provided by the SEXPE service. 
\item Spanish National Statistics Institute (INE)\footnote{https://cutt.ly/gs0gzeP}. This agency is responsible for publishing national statistics about different domains in the country. In this study we obtained the list of contract economical sectors from this source.
\end{itemize}

\subsection{Data gathering}

Based on the four data sources commented in previous section, the data owners provided us with the VET student's records stored in Rayuela for citizens since 1996. These data were also completed, on the one hand, with the job contracts stored by the SEXPE service for the citizens included in the previous dataset (the provided by Rayuela) and, on the other hand, the last information provided by the National Social Security Agency for these citizens (note that in both cases the information included the whole life for each citizen). The data provided by INE were also useful for completing the information available for each citizen. 

Based on the different data sources, the files used in the study are summarized in Table \ref{table:DataSourcesDescription} together with the number of instances, features and a description for each one.

\begin{table*}[!htbp]
{
\centering
{\fontsize{8.5pt}{9.5pt}\selectfont
\begin{tabular}{l  p{1.5cm}  r  r  p{5cm}l}
	\hline
	{\bf File} 	& 	{\bf Source} & 	{\bf Features} & {\bf Instances} & 	{\bf Description}   \\
	\hline		
   Graduates 	& Rayuela & 14  & 28.272 & Graduates data from VET Schools in Extremadura from 1996 to 2016.\\ 	
   \hline	
   Contracts  	& SEXPE & 16  & 317.152 & New employees contracts including the sector, economic area and location from 1990 to 2016.\\
   \hline
   SocSec  	& Soc. Sec. Agency 		  & 14  & 216.726 & Complementary information about citizen labor periods. The range of dates is from 1984 to 2016 \\	 \hline 	
   CNAE  	& INE 		  & 5  & 993 & Spanish National Classification of Economic Activities that encode companies economic areas.\\	
	\hline
\end{tabular}
}
\caption{Datasets Description}
\label{table:DataSourcesDescription}
}
\end{table*}

\subsection{Data warehouse design}

\begin{figure}[htpb]
\centering
\includegraphics[width = .8\linewidth]{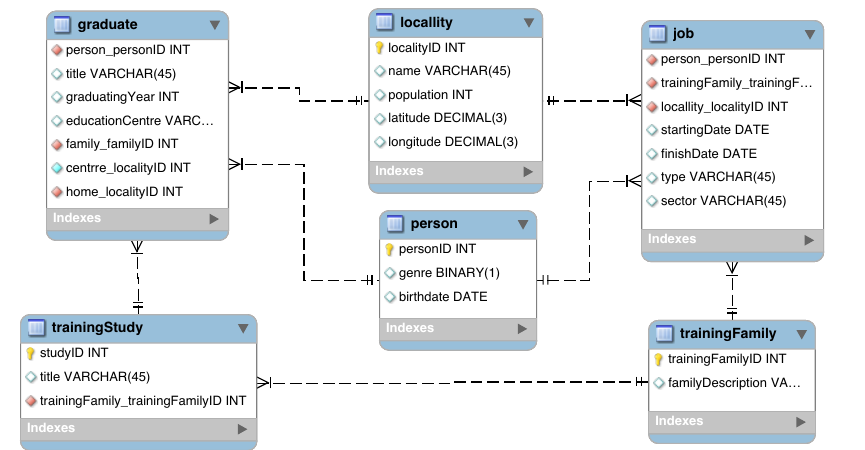}
\caption{Data Warehouse Schema.}
\label{fig:dataWarehouseSchema} 
\end{figure}

In order to be able to query and process all the datasets available, we needed to design a data warehouse so that all this information could be integrated into the same storage. To properly organize the data we used the typical \textit{Snowflake Schema} \citep{datawarehousingORACLE} logical design, since the structure of our data perfectly fits with this type of design. Note that most of the datasets contained the personal data of the citizen as key identifier. Thus, the design contains a central table (person) with this information that makes as pivot table to connect to the rest of them. The data warehouse designed is illustrated in Figure \ref{fig:dataWarehouseSchema}.

Observe that \textit{person} table plays the role of \textit{fact table} in our design, whilst \textit{graduate} and \textit{job} act as \textit{dimension tables}. Additionally, based on the normalization performed, \textit{locality}, \textit{trainingStudy} and \textit{trainingFamily} tables are added as dimension tables of \textit{graduate} and \textit{job} ones. \textit{Locality} stores the location of the family and education centre, \textit{trainingStudy} stores the list of different VET programs included in our analysis and \textit{trainingFamily} stores the list of VET professional families. These additional dimension tables are the reason why we used a snowflake design instead of a star one. 

These tables were built based on the datasets explained in the previous section. Obviously, as it may be observed in Figure \ref{fig:dataWarehouseSchema}, there are some datasets that are not reflected in the design. Concretely, \textit{SocSec} dataset was used to complete the information included in \textit{job} table whilst \textit{CNAE} dataset, which contains the relationships among VET professional families and job economical sectors, was used to derive the relation between \textit{job} and \textit{trainingFamily} tables.

\section{Methodology}
\subsection{TOPSIS}
The Technique for Order of Preference by Similarity to Ideal Solution (TOPSIS) is a non-parametric multi-criteria ranking method which is model-free and data-driven. Following the notation of \cite{Haang2011}, let us assume we have $m$ alternatives $A_1,\ldots,A_m$ and $n$ criteria $C_1,\ldots,C_n$, with weights $\omega_1,\ldots,\omega_n$. Each criterion may be either a benefit (i.e. \textit{``the more the better''}) or a cost (i.e.  \textit{``the less the better''}). Let $J^+$ and $J^-$ denote the sets of indices $j$ corresponding to the benefit and cost criteria, respectively, and let $X$ denote the $m\times n$ matrix of performance ratings of each alternative at each criterion. 

The TOPSIS algorithm encompasses the following steps: 
\begin{enumerate}
\item Normalize and weight the columns of $X$, obtaining the normalized matrix $N$ whose element $ij$ is defined by
\[
n_{ij} = \omega_j\frac{x_{ij}}{\sqrt{\sum_{k = 1}^m x_{kj}^2}},\quad i = 1,\ldots, m;\,j= 1,\ldots,n.
\]
\item Find the ideal, $A^b$, and antiideal $A^w$ solutions:
%\begin{align*}
\[
A^b_j = \begin{cases}
\displaystyle{\max_{i = 1,\ldots, m}}\,n_{ij}&\text{if } j \in J^+\\
\displaystyle{\min_{i = 1,\ldots, m}}\,n_{ij}&\text{if } j \in J^-
\end{cases}\quad
A^w_j = \begin{cases}
\displaystyle{\min_{i = 1,\ldots, m}}\,n_{ij}&\text{if } j \in J^+\\
\displaystyle{\max_{i = 1,\ldots, m}}\,n_{ij}&\text{if } j \in J^-
\end{cases}
\]
%\end{align*}
\item Find the distances between each alternative and the ideal and antiideal solutions:
%\begin{align*}
\[
d_i^b = \|A_i-A^b\| = \sqrt{\sum_{j = 1}^n (n_{ij} - A^b_j)^2}\,;\quad
d_i^w = \|A_i-A^w\| = \sqrt{\sum_{j = 1}^n (n_{ij} - A^w_j)^2}
\]
%\end{align*}
\item Compute the score ranking for each alternative $r_i = \frac{d_i^w}{d_i^w+d_i^b}$.
\item Sort the alternatives according to the scores $r_i$. 
\end{enumerate}

This method is very easily implemented and applied, however, as usual, devil is in the details and in this case, the weight determination is a major concern since  a slight change in them can lead to different final rankings.

\subsection{Criteria definition and weight determination}
In order to obtain the different rankings presented in this work, first the criteria that would allow us to compare the degree of employability of the different VET programs must be determined. A Delphi methodology was used for defining such criteria and their corresponding weights \citep{dalkey63}. In particular, 10 experts were asked in rounds to define the main criteria  that would take into account both, the employability when the VET program is related with the job economic sector, and the employability in general, i.e., regardless the job economic sector and the family of the VET program studied. 
After reaching a consensus, the criteria defined were:

\begin{enumerate}
\item[\textbf{C}$_1$] Time elapsed from the graduation to the signing of the first employment contract within the same professional field as the VET program. 
\item[\textbf{C}$_2$] Average time elapsed between the end of a contract and the signing of the next employment contract within the same professional field as the VET program.
\item[\textbf{C}$_3$] Fraction of the labor life worked within the same professional field as the VET program. 
\item[\textbf{C}$_4$] Fraction of the time worked within the same professional field as the VET program under temporary contracts.
\item[\textbf{C}$_5$] Time elapsed from the graduation to the signing of the first employment contract regardless the professional field.
\item[\textbf{C}$_6$] Average time elapsed between the end of a contract and the signing of the next employment contract regardless the professional field.
\item[\textbf{C}$_7$] Fraction of the labor life under temporary contracts.
\item[\textbf{C}$_8$] Time spent without contracts.
\end{enumerate}

Criteria \textbf{C}$_1$ - \textbf{C}$_4$ analyse the employability of a VET program in its corresponding professional field. On the other hand, criteria \textbf{C}$_5$ - \textbf{C}$_7$ analyse the employability of a VET program in general, regardless of its professional family. The reason for including the latter is because it was found that for many degrees, most of the graduates did not get to work within the scope of their professional family, but still they managed reasonably well, to found a job in other fields and somehow this indicated that for some professions, the fact of just finishing a VET Program was enough. Finally, criterion \textbf{C}$_8$  is meant to penalize long term unemployment.

Regarding the criteria weights, the experts were asked to set the minimum weight to 1 and then to quantify the rest of the weights in relative importance to that of the minimum. Afterwards some discussion was conducted among the experts until a consensus was reached. The values obtained are given in Table \ref{tab:weights}. Those weights clearly reflect the importance given to the employability of the different VET programs within their professional family. 

\begin{table}
\centering
\begin{tabular}{ccccccccc}
&\textbf{C}$_1$&\textbf{C}$_2$&\textbf{C}$_3$&\textbf{C}$_4$&\textbf{C}$_5$&\textbf{C}$_6$&\textbf{C}$_7$&\textbf{C}$_8$\\\midrule
$W_i$&4&2.5&1&1&3&2&1&1\\
$\omega_i$&$0.258$&$0.161$&$0.065$&$0.065$&$0.193$&$0.128$&$.065$&$0.065$\\
\bottomrule
\end{tabular}
\caption{Relative and absolute weights of the criteria determined by the Delphi methodology.}
\label{tab:weights}
\end{table}

\section{Criteria influence determination}

As was mentioned earlier, one of the most delicate parts in the design of a ranking score is the weighting of each of the components that define it. The use of expert opinion is common when determining those weights, however, the heterocedasticity and the correlation between the different criteria make these assigned weights rarely coincide with their real influence in the final score ranking \citep{Paruolo2013}.

%Even though the DELPHI method is a widely used method for determining both criteria and their corresponding desired weights, within the framework of composite indicators as the one presented here, it is critical to assess the influence of the different criteria in the final outcome of the ranking procedure.
In order to assess this influence, we are going to use two different approaches. The first one is the \textit{sensitivity analysis} \citep{Paruolo2013} which is based in the Pearson's correlation ratio, while the second one, which we introduce in this work, is based on the assessment of the performance under different scenarios. 

\subsection{Global sensitivity analysis based on the use of variance}
The sensitivity analysis technique used here, is applied to assess the quality of the ranking score by finding  the ``\textit{effective weights}'' of the criteria (main effects), providing thus a measure of the real importance of each criterion.

The sensitivity analysis method determines the actual contribution of each criterion to the overall ranking by means of the Karl Pearson's correlation ratio, $\eta^2$ which, for each criterion $C_j$, is defined as \citep{saltelli2010}
\begin{equation}
\label{eq:sens}
\eta^2_j = \frac{\text{Var}_{X_j}(E_{X_{\textrm{-}j}}(r|X_j))}{\text{Var}(r)},     
\end{equation}
where $X_j$ denotes the $j$-th column of the performance matrix $X$ (i.e. the vector of the performance of all $m$ alternatives in the $j$-th criterion), $X_{\textrm{-}j}$ denotes the vector containing all except the $j$-th input of vector $X_j$, and $r$ denotes the vector of the final scores of all $m$ alternatives. Parameter $\eta_j^2$ is known as the \textit{main effect} of criterion $C_j$ on the output vector $r$, in terms of the expected output variance explained  by $X_j$ \citep{Bas2017}. The greatest difficulty in calculating these $\eta_j^2$ lies in obtaining the numerator in \eqref{eq:sens} which, in turn, depends on the conditional expectation of $r$, given $X_j$. To overcome this problem we use the \textit{``dependent regression''} method \citep{ratto2007},
which mainly consists on an non-parametric smoothing algorithm based on a Kalman filter.

\subsection{Comparing scenarios}

The different scenario comparison we introduce here is inspired in the Kao-Liu fuzzy DEA method of best-worst case scenarios \citep{Kao2003}. For a given criterion $C_j$ we will consider two possible scenarios: 
\begin{itemize}
\item Most weighted case: in this case we will assume that $C_j$ has  the highest relative weight among all criteria. As a reference measure we assign the relative weights as $W_j^+ = 2$, $W_i^+ = 1$ for $i \not= j$, i.e. in the most weighted scenario, criterion $C_j$ is assigned twice as much importance as the rest of weights. Therefore, in our case, the absolute weights in this case will be $\omega^+_j = 0.22$, $\omega^+_i = 0.11$ for $i\not=j$.
\item Least weighted case: in this case, criterion $C_j$ has half the weight of the rest of criteria, i.e., $W_j^- = 1$, $W_i^- = 2$ for $i \not= j$. Hence, the weights used for this case in our study will be $\omega^+_j = 0.066$, $\omega^+_i = 0.133$ for $i\not=j$.
\end{itemize}

Thus, for each criterion we obtain two rankings $R_j^+$, and $R_j^-$ which are represented by two permutations of the numbers $1,\ldots, m$. The identity permutation is assigned to the first ranking, and the permutation for the second ranking is obtained using the first as a reference. 

Finally, the relative Kendall - tau distance between both permutations, $D_j$ is computed. This distance measures the fraction of all possible pairings of alternatives which change the order from one ranking to the other.

If this value is small, it can be interpreted  as a lack of influence of criterion $j$ in the preference ordering of the alternatives, since no matter how big or small the weight of the criterion is, the final ranking is not varying that much.

\section{Results and discussion}

\subsection{Data preparation}

The original data set consisted on a total of 28.272 people graduating in 121 different VET programs from 1996 to 2016. Those VET programs were classified according to the professional family they are included in. For each VET program and for each graduating year the scores of each person verifying the conditions of the different criteria were obtained. Then the median of such scores was calculated and considered as the score of that program for that year in the corresponding criterion.

\begin{figure}[H]
\centering
\includegraphics[width = .75\linewidth]{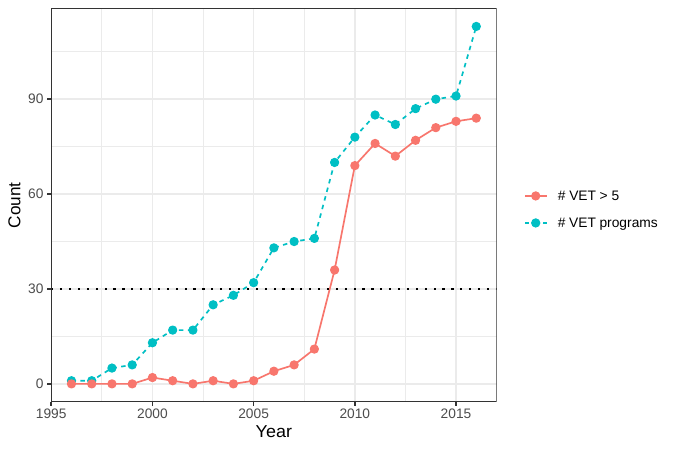}\vspace{-0.5cm}
\caption{Number of total VET programs available each year (dashed blue line) and number of VET programs with all criteria computed as medians of more then 5 data (solid red line). The horizontal dashed line shows the minimum number of programs considered for the ranking.}
\label{fig:nvetprograms} 
\end{figure}

Nevertheless, some filtering had to be carried out previously. First, not all programs had scores in all criteria all years.  Indeed, for the first years in the series, there is a very small number of VET programs scoring in all criteria. Thus we only considered the years for which we had a reasonable amount of programs with data in all criteria. We set as threshold of 30 VET programs as the minimum for making the ranking. 

Secondly, as previously stated, the scores in the different criteria are obtained as a median of the scores of individual persons. However, in some cases, such median was computed over very few people which could lead to biased  results. Therefore we only considered VET programs for which the scores were obtained from more than 5 data. 

Figure \ref{fig:nvetprograms} depicts both, the number of VET programs available for each year (from 1996 to 2016) and the number of VET programs for which all the criteria were obtained from data from more than 5 people. According to the number of programs in each year, we considered the time window from 2009 to 2016.  

After the filtering, from the original 121 VET programs, a total amount of 106 were used in this study\footnote{The list of all the VET names and Professional families is available in https://github.com/chemacmunex/vetrankings.git}. 

%See Table \ref{tab:vetfamilies} and Table \ref{tab:vetfamiliescourses} in the supplementary information (appendix) for the name of the VET programs and their corresponding professional families. 

\begin{figure}%[H]
\centering
\includegraphics[width = \linewidth]{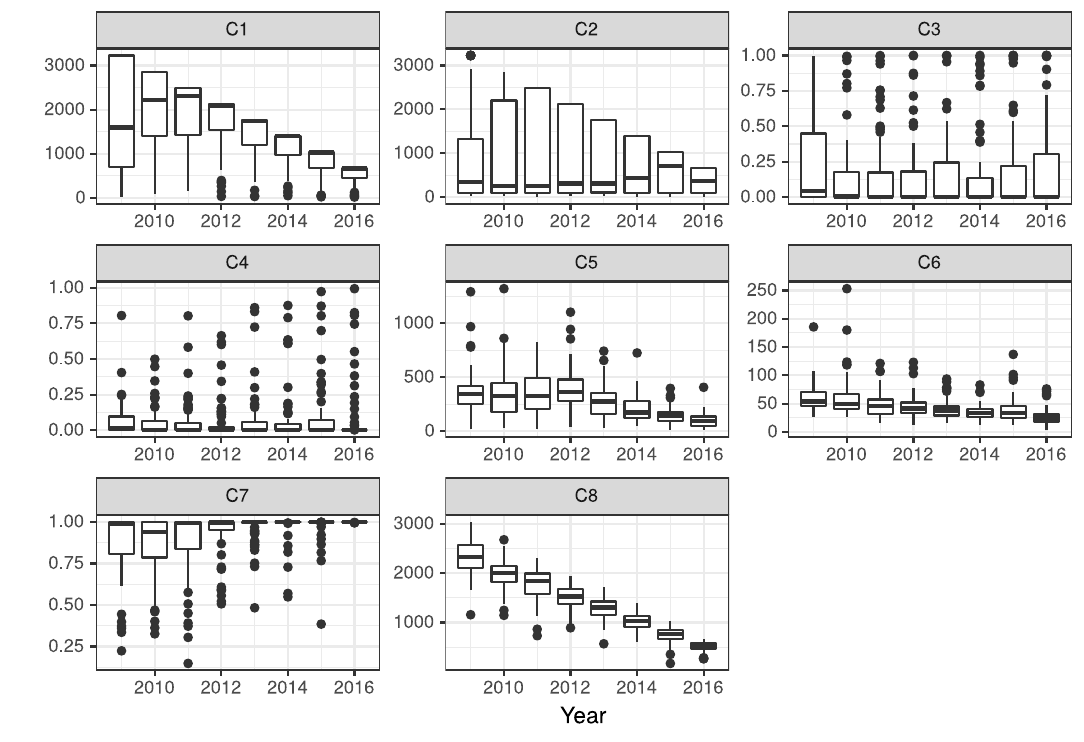}\vspace{-1cm}
\caption{Evolution of the general performance at each criterion in the period 2009-2016.}
\label{fig:critevol} 
\end{figure}

As a brief descriptive summary of the data, Figure \ref{fig:critevol} shows the evolution of the performance scores of each criterion throughout the period 2009 - 2016. 

The general behaviour of the criteria reflects the evolution of the economy in Spain, particularly in Extremadura. The time required to find the first job (criteria C1 and C5)  starts a clearly decreasing trend in 2012, together with the time between contracts (criteria C2 and C6). For the rest of the criteria there is no general improvement (in median). This turning point in 2012 coincides with the moment when the region's GDP growth bottomed out and began to recover, although it was not until 2014 when it began to have positive values, like the rest of the country.

\subsection{VET rankings}
\begin{figure}[htpb]
\centering
\includegraphics[width=\linewidth]{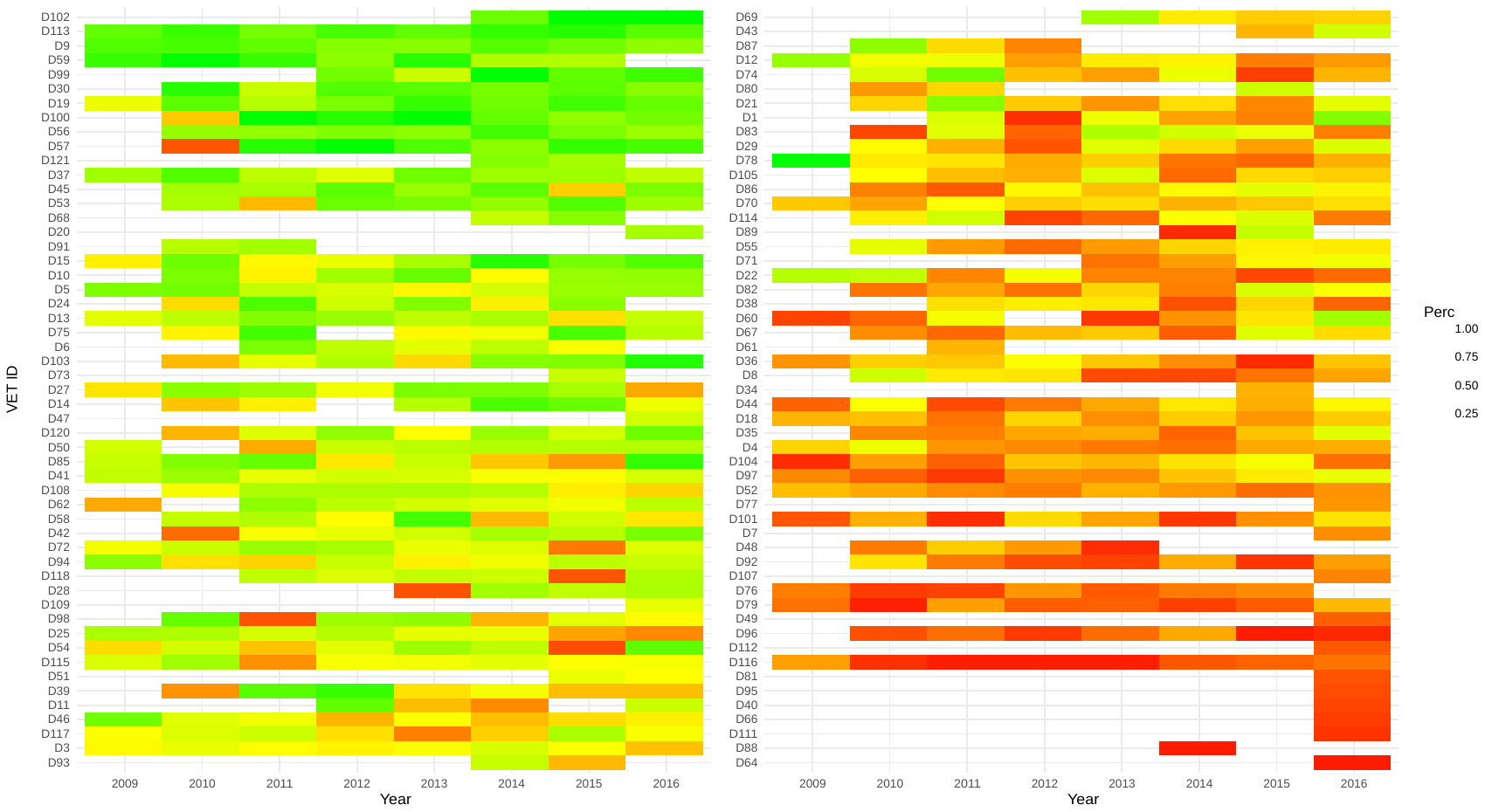}\vspace{-0.5cm}
\caption{VET programs ranking throughout the period 2009-2016. 	
The colour reflects the percentile of each program at each year (the greener the better) and the overall sorting has been done according to the average percentile of each program during the whole time span.}
\label{fig:vetrks}
\end{figure}

Figure \ref{fig:vetrks} shows the results for the rankings of the VET programs from 2009 to 2016.  	
The figure shows, with a colour, the percentile of each VET program at each year. The VET programs have been sorted according to their average percentile during the whole period 2009-2016.
This plot allows to easily find both, the overall performance of a program and the evolution over time of such performance. 

A closer look at the top and bottom of the ranking, as shown in Table \ref{tab:bestworst}, gives us some insight about which VET programs are the best behaved regarding the employability. Seven out of the 10 best programs are from the professional families ``\textit{Agricultural and Livestock Activities}'' (ALA) and ``\textit{Hostel and Tourism}'' (HOT), which correspond with the two major economic activities (or sectors) in the region: Primary and Tertiary sectors, namely agriculture and tourism. 

\begin{table}[H]
\begin{tabular}{cllcr}
\toprule
Pos.&ID & VET Program & Mean P. &Pr. F.\\
\midrule
1&D102 & Prod. Programming in Mech. Manufa. & 0.968 & MEM \\
2& D113 & Catering Services & 0.944 & HOT \\
3& D9 & Physical and Sport Activities Org. & 0.898 & PSA \\
4& D59 & Oral hygiene & 0.894 & HEA \\
5& D99 & Agroecological Production & 0.888 & ALA \\
6& D30 & Cookery Management & 0.885 & HOT \\
7& D19 & Cooking and gastronomy & 0.855 & HOT \\
8&D100 & Agricultural and Livestock production & 0.853 & ALA \\
9&D56 & Forest and Natural Envmnt. Mgmt. & 0.852 & ALA \\
10&D57 & Mgmt. \& Org. of Agr. \& Livestock Co. & 0.826 & ALA \\
%11&D121 & Viticulture & 0.817 & FOI \\
%12&D37 & Kindergarden & 0.778 & CSS \\
%13&D45 & Aesthetics & 0.778& PIM \\
%14&D53 & Tourist accom. Mng.& 0.773 & HOT\\
%15&D68 & Heater instal. & 0.768 & IMA \\
\vdots&\vdots&\vdots&\vdots&\vdots\\
%92&  D92 & Bakery and Confec. & 0.180 & FOI\\
%93&  D107 & Radiotherapy & 0.178 & HEA\\
%94& D76 & Lab. & 0.131 & CHE\\
%95&D79 & Imag. Lab. & 0.131 & IMS\\
%96&  D49 & Man. and assembly & 0.100 & MEM\\
97&  D96 & Prepress in Graphic Arts & 0.096 & GRA\\
98&D112 & Commercial services & 0.089 & COM\\
99&D116 & Microcomputer systems and networks & 0.081 & ITC\\
100&  D81 & Vehicle maintenance & 0.078 & VTM\\
101&D95 & Hairdressing and aesthetics & 0.067 & PIM\\
102&D40 & Electricity and electronics & 0.056 & ELE\\
103&  D66 & IT and communications & 0.044 & ITC\\
104& D111 & Administrative services & 0.033 & ADM\\
105&D88 & Tesellations & 0.012 & ART\\
106&D64 & Food industries & 0.011 & FOI\\
\bottomrule
\end{tabular}
\caption{Head and tail of the ranking. Position, mean percentile throughout the period 2009-2016 and the corresponding professional family is tabulated for the first and last 10 VET programs.}\label{tab:bestworst}
\end{table}

\begin{figure}
\centering
\includegraphics[width = \linewidth]{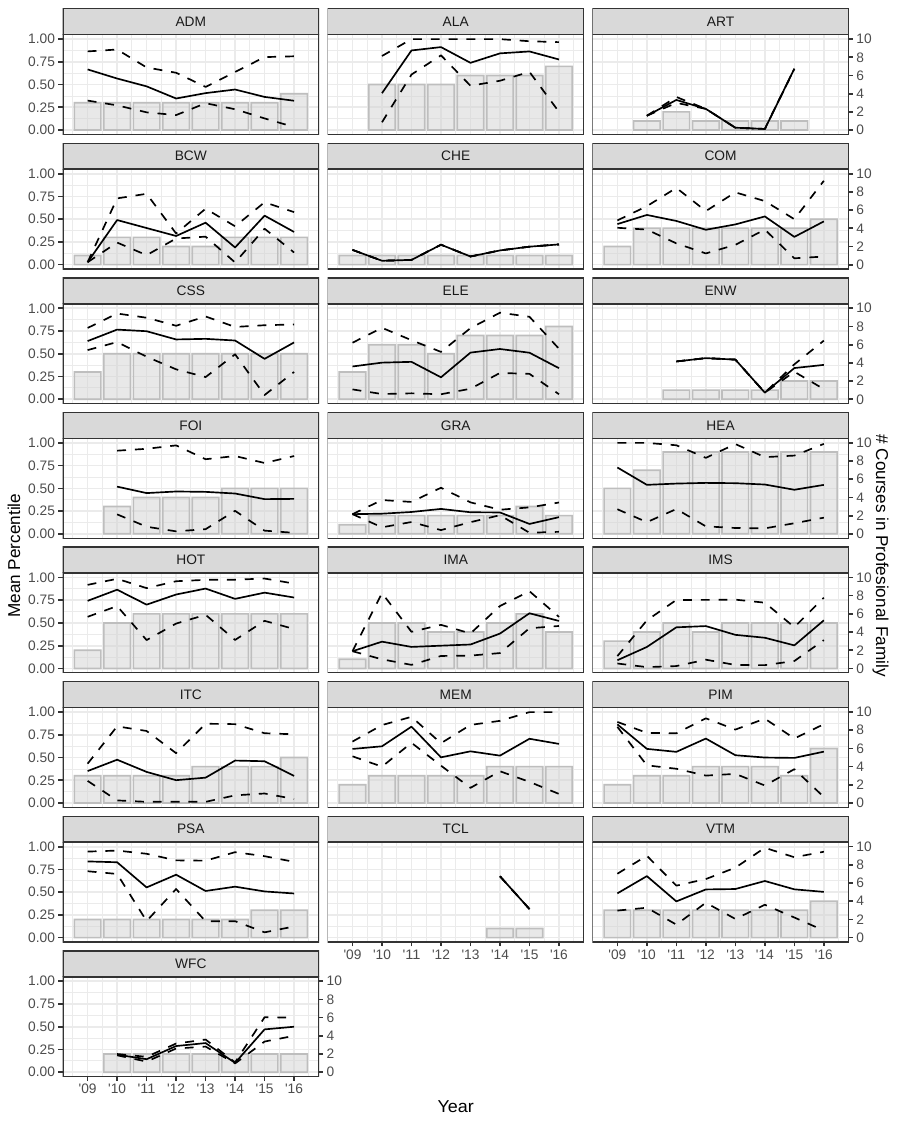}\vspace{-1cm}
\caption{ Mean percentile of the VET programs in each professional families (solid line), minimum and maximum percentiles (dashed lines) and number of VET programs considered each year (gray bars).}\label{fig:proffam}
\end{figure}

It is also interesting to analyse how the programs within each professional family are doing regarding the employability of their students. Figure \ref{fig:proffam} depicts, for each professional family, the evolution over time of their programs in the ranking. We show the mean of the percentiles obtained by the VET programs of the corresponding professional family (solid lines) together with the minimum and maximum values (dashed lines) and the amount of programs that each year fulfilled all the conditions mentioned above, for being considered in the ranking (gray bars - right side y-axis). 

It is clearly seen how Agriculture and Livestock Activities (ALA) and Hostel and Tourism (HOT) perform very well through most of the period with very few differences between the maximum and the minimum. Other professional families, on the contrary, perform poorly for most of the time, like Graphic Arts (GRA), Chemistry (CHE) and Arts and Artcraft (ART). 

Finally, some other professional families, like, for example, Administration and Management (ADA), Health (HEA), IT and communications (ITC) or Personal Image (PIM),  present huge differences between the maximum and the minimum. In some cases those differences can be explained because the programs within the same professional families are very different in nature: for example, in the Health professional family, Dentures (D103) and Oral Hygiene (D59) programs are ranked in the top of the list while Anatomical Pathology and Cytodiagnosis (D8), Radiotherapy (D107) or Clinical diagnostic laboratory (D78) perform quite poorly in comparison. These professions differ both in the workplaces where they take place and in the activity itself.

However, in other cases, such as the Personal Image professional family, all programs, with the exception of Personal and Corporate Image Consulting VET program, are related to hairdressing and aesthetics, but their positions in the ranking range from as low as a 6.6\% for Hairdressing and Aesthetics (D95) to an average percentile of 77\% for Aesthetics (D45) programs. This high variability may be due to an over fragmentation of the professional family in too many programs with very similar characteristics.

\subsection{Criteria influence analysis}

\begin{figure}
\centering
\includegraphics[width = 0.75\linewidth]{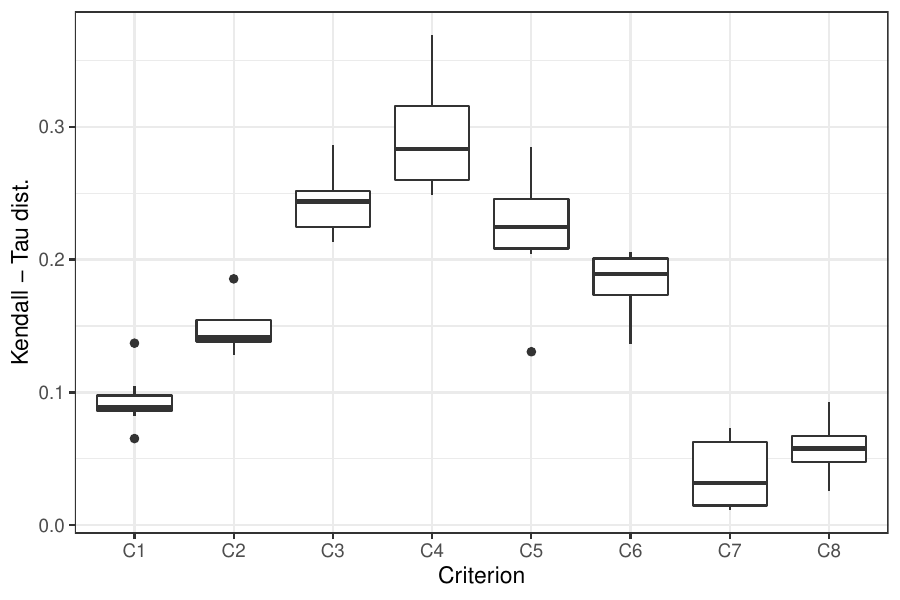}\vspace{-0.5cm}
\caption{Kendall - tau distance between the least and most weighted scenarios for each criterion.}\label{fig:scenario}
\end{figure}

Figure \ref{fig:scenario} depicts a boxplot of the Kendall - tau distance between the most weighted and the least weighted scenarios for each criterion for the whole 2009-2016 period, showing firstly that criteria C7 (Fraction of the labor life under temporary contracts) and C8 (Time spent without contracts) are the least sensitive to weight changes with a median Kendall - tau distance of around a 5\% and secondly, that criterion C4 (Fraction of the time worked within the same professional field as the VET program under temporary contracts) is the most sensitive to those weights changes (with a Kendall - tau distance reaching values larger than 35\%).

\begin{figure}
\centering
\includegraphics[width = \linewidth]{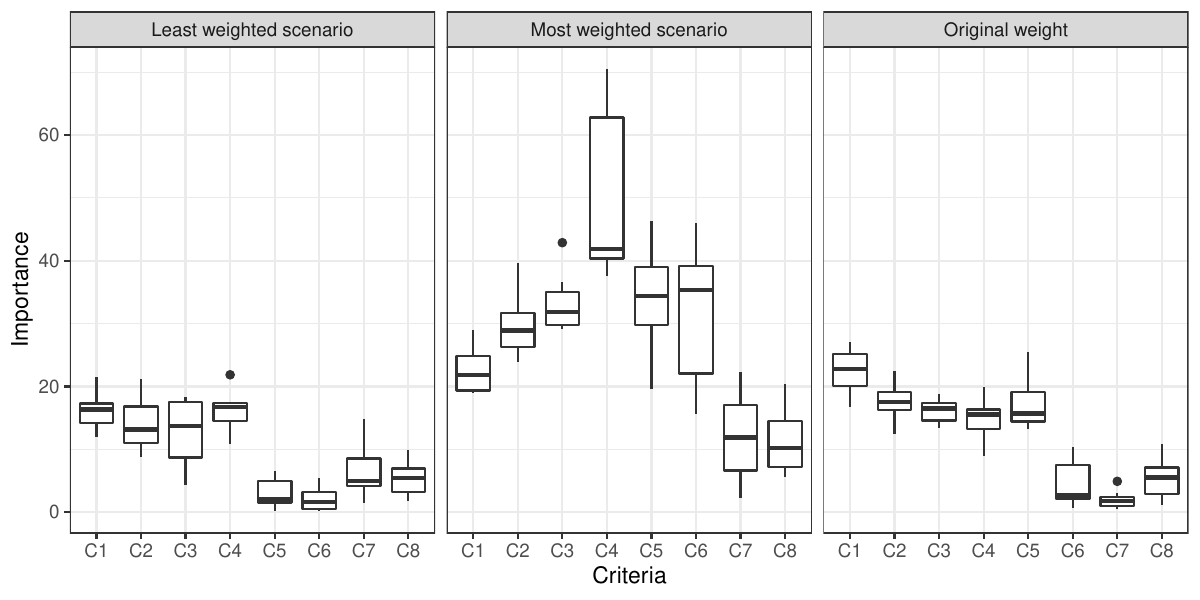}\vspace{-1.1cm}
\caption{ Global sensitivity analysis for the three cases considered: the least weighted scenario (left), the most weighted scenario (middle) and the ranking obtained with the original weights given by the expert panel (right).}\label{fig:sensitivity}
\end{figure}

On the other hand, we can compare these results with those obtained from the Global Sensitivity Analysis. Figure \ref{fig:sensitivity} depicts the Karl Pearson's correlation ratio for three different rankings: the least weighted scenario (left plot), the most weighted scenario (middle plot) and the weights proposed by the expert panel (right plot). Taking a closer look to the three plots we can see that the importance of  criteria C7 and C8 in the construction of the final score is almost the same and it is very low. The importance of criteria C1 and C2 show more differences across the three rankings, and finally, criteria C3, C4, C5, and C6, show the largest differences in importance between the three rankings, showing a larger importance when the weights assigned are larger and lower values of importance when their assigned weights are smaller.  
This agrees with the results of the scenario comparison method that showed that these criteria were very sensitive to changes in the assigned weights.

However, the scenario comparison method introduced in this work, focuses not on the scores obtained by the MCDM method (TOPSIS score in this case), like the Global Sensitivity index does, but on the final ranking obtained. This is important since a change in the score of an alternative does not necessarily lead to a change in its position in the ranking. Also it is worth mentioning that the scenario comparison method introduced here has the advantage of not needing a previous definition of the criteria weights, and therefore the information about the sensitivity of the final ranking to the weight assigned to each criterion is available to the expert panel before they make the weight assignment.  

Moreover, if we recall the weights assigned by the expert panel, it may look that there is a divergence between the weights assigned to each criterion and their sensitivity, as was found in the previous results. This divergence does not mean that the weights were wrongly assigned by the experts. In this case the experts (regional administration) gave more importance to criteria C1 and C2, mainly because those criteria were directly related to  finding a first job rapidly and keep working on a labour sector directly related to the study field of the VET, and this could encourage young people to enrol on such VET programs. On the other hand criterion C4 was regarded as a long term criterion which was considered to be less important for young people when deciding whether to take a VET course or not. 

Notwithstanding the above, this tool does give information about the changes that the final ranking may have if, for political reasons, it is decided to focus the importance of the criteria in a slightly different way. In particular, the experts should know that a  slight change is made in the weight of criterion C4, the final ranking will be far more affected than if this change is made to criteria C1 or C2 (for example). With this information, experts could establish the allocation of weights in a safer way and with information consistent with the allocation.

Therefore, to some extent, the proposed criteria analysis is independent from the weight definition, but decision-makers should be aware of the sensitivity of the different criteria to changes in the weights, since those changes can be determined by changes in some policies.

%These results partly agree with the Global sensitivity analysis shown in Figure \ref{fig:sensitivity}.  The Karl Pearson's correlation ratio has been computed for three different rankings: the least weighted scenario (left plot), the most weighted scenario (middle plot) and the weights proposed by the expert panel (right plot). In all of them criteria C7 and C8 were the ones with the least influence on the final ranking scores, result which agrees with the aforementioned Kendall-tau distance. On the other hand, in the most weighted scenario (Figure \ref{fig:sensitivity}, middle plot), C4 is the criterion which influences the final ranking score the most, in a similar fashion to the result obtained with the Kendall - tau distance technique.
%The lack of sensitivity to weight changes as well as effective influence of criteria C7 and C8 seems to point in the direction that the most important criteria are those related to the employability within the same professional field as the VET program studied. Criteria C7 and C8 do not fall into this category. However, criterion C6, which does not take into account the professional field of the contract either, also shows a small effective influence (at least in two of the three scenarios considered), but it shows a relatively important dependence on the weight determination, as shown by the Kendall-tau distance. 
%This supports the idea that  the criteria weight determination must be chosen very carefully since they may affect enormously to the final ranking even for the non-influent criteria. 

\section{Conclusions}
%Extraído del Report original
%En este trabajo se ha probado la efectividad del método TOPSIS como herramienta de análisis multicriterio para la clasiﬁcación de títulos de formación profesional en Extremadura durante el periodo 2009-2016.
In this work we have proven the effectiveness of the TOPSIS method as a multi-criteria analysis tool for the classification of VET programs in Extremadura during the period 2009-2016.  	
%Es muy importante recalcar el hecho de que tener al alcance datos cuantitativos que relacionan distintas bases de datos de las Consejerías de Educación y Trabajo permite tener más criterios en los que basar los ránkings. Hay que tener en cuenta que normalmente o bien los rankings son cualitativos y están basados en encuestas de satisfacción sobre su situación laboral una serie de años después de haber terminado los estudios (como por ejemplo el proyecto REFLEX [3] en el ámbito de la educación superior) o, en el caso de los rankings cuantitativos están, en la mayoría de los casos, basados en un único criterio: porcentaje de egresados que está trabajando al año de haber terminado los estudios (por ejemplo).
It is very important to emphasize the fact that having quantitative data available relating different databases from the Education and Employment Board of Extremadura, allows us to have more criteria on which to base the performance of the different VET programs. It should be borne in mind that in this type of rankings, the criteria employed are usually either qualitative and based on surveys about the satisfaction  on their employment status several years after completing their studies or, in the case of quantitative criteria, they are, in most cases, based on a single criterion (e.g. percentage of graduates who are working a year after completing their studies).

In the analysis, we have provided a classification considering 8 different criteria, that were calculated based on the real data provided by the Education and Employment Board of Extremadura. This is a significant contribution of this work with respect to similar ones that apply multi-criteria analysis relying on the use of questionnaires. 

In addition, in this work we have proposed a new decision support method for assessing the influence of the weights assigned to the different criteria in the final ranking, comparing it with other known technique for criteria influence analysis. This new method can be applied before the weight definition by the experts, so it can serve as support information for the determination of the final weight scheme. This analysis showed that it is of paramount importance a thorough analysis of the criteria, since the ranking of the different VET programs, and all the possible policy decisions based on such ranking, greatly depend on the choice of the weight assigned to each criterion. 

Finally, we strongly believe that the approach presented in this work, in which the academic data of each VET student is linked to its full labor history, is a step forward in the analysis of the efficiency and usefulness of such VET programs. Moreover, since VET programs are widely consolidated all over Europe, this approach could be easily replicated in other countries and regions, using their own citizens' data, to try to reach the same goal.
%Moreover, since this is novelty approach to study the relationships between VET programs and labor market and VET programs are widely consolidated all over Europe, the approach could be easily replicated in other countries and regions, using their own citizens' data, to try to reach the same goal.

\section*{Acknowledgements}
 This  work  has  been  developed  with the support of (i) Ministerio de Ciencia e Innovaci\'on (MCI), Agencia Estatal de Investigaci\'on (AEI) and European  Regional  Development  Fund  (ERDF): RTI2018-098652-B-I00 and RTI2018-093608-B-C33 projects, and (ii) European  Regional Development Fund (ERDF) and Junta de Extremadura: IB16055, IB18034 and GR18112 projects.

\section*{References}

\bibliography{references.bib}
\bibliographystyle{abbrvnat}

\end{document}